\documentclass[letterpaper, 10 pt, conference]{ieeeconf}  
\IEEEoverridecommandlockouts
\usepackage{stmaryrd}
\usepackage{amssymb}
\usepackage{amsmath}
\usepackage{graphicx}
\usepackage{mathtools} 
\usepackage{booktabs}
\usepackage{multirow}
\usepackage{xcolor}
\usepackage{cleveref}
\usepackage{algorithm}
\usepackage{algorithmic}
\usepackage{subcaption}
\usepackage[utf8]{inputenc}
\usepackage[T1]{fontenc}

\usepackage[natbib=true,
            style=numeric,
            sorting=none,
            backend=biber,
            maxcitenames=6,
            maxnames=6
            ]{biblatex}
            \AtBeginBibliography{\small}

\newcommand{\algorithmiccontinue}{\textbf{continue}}
\newcommand{\CONTINUE}{\STATE \algorithmiccontinue}

\DeclareMathOperator*{\argmin}{argmin} 

\title{\LARGE \bf
Lidar-Monocular Surface Reconstruction Using Line Segments
}

\author{Victor Amblard$^1$, Timothy P. Osedach$^2$, Arnaud Croux$^3$,
Andrew Speck$^3$ and John J. Leonard$^4$
    \thanks{This work was supported by Schlumberger Technology Corporation
and ONR Grants
N00014-18-1-2832 and N00014-19-1-2571.}
    \thanks{$^1$Currently at Mines ParisTech; work performed at MIT CSAIL;
Email:  {\tt victor.amblard@mines-paristech.fr}}
    \thanks{$^2$Work performed at
Schlumberger Doll-Research; Current affiliation is General Dynamics Mission Systems; Email:
    {\tt tim.osedach@gmail.com}}
    \thanks{$^3$Schlumberger-Doll Research, Cambridge MA; Email:
{\tt ACroux,ASpeck@slb.com}}
    \thanks{$^4$MIT CSAIL; Email: {\tt jleonard@mit.edu}}}

\begin{document}

\maketitle


\begin{abstract}
  Structure from Motion (SfM) often fails to estimate accurate poses
in environments that lack suitable visual features. \
In such cases, the quality of the final 3D mesh, which is contingent on
the accuracy of those estimates, is reduced.  One way to
 overcome this problem is to combine data from a monocular camera with
that of a LIDAR. This allows fine details and texture to
 be captured while still accurately representing featureless subjects. However, fusing these
two sensor modalities is challenging due to their fundamentally
different characteristics. Rather than directly fusing image
features and LIDAR points, we propose to leverage common geometric
features that are detected in both the LIDAR scans and image 
data, allowing data from the two sensors to be processed in a
higher-level space. In particular, we propose to find correspondences
 between 3D lines extracted from LIDAR scans and 2D lines detected
in images before performing a bundle adjustment to refine poses. We also exploit the
detected and optimized line segments to improve the quality of the final
 mesh. We test our approach on the recently published
dataset, Newer College Dataset. We compare the accuracy
 and the completeness of the 3D mesh to a ground truth obtained with a
survey-grade 3D scanner.  We show that our method delivers results that are 
 comparable to a state-of-the-art LIDAR survey while not requiring
 highly accurate ground truth pose estimates. 
\end{abstract}

\section{Introduction}
Dense reconstruction of large scenes such as buildings or outdoor environments are increasingly compelling for a variety of applications including inspection, change detection and 3D asset creation. Survey-grade 3D LIDAR systems are able to produce very accurate reconstructions of environments but are expensive and can be time-consuming to comprehensively cover large areas. 
On the other hand, reconstructing a 3D scene solely from monocular images presents a different set of challenges.
First, a large number of images with significant overlap must be collected for the poses to be accurately recovered. Additionally, the scene must be well-lit and sufficiently rich in visual features.
Finally the scene that is reconstructed may not be referenced to an absolute scale.
\begin{figure}[!h]
  \centering
  \begin{subfigure}[b]{0.475\linewidth}
      \centering
      \includegraphics[width=\textwidth]{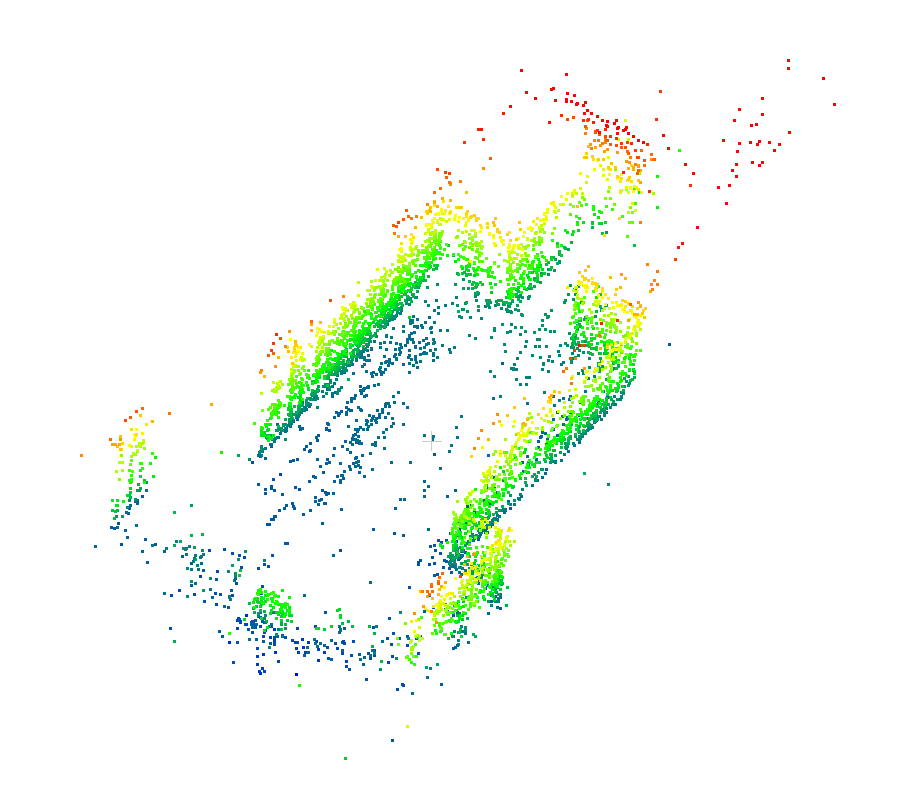}
      \caption[Network2]%
      {{\small Visual features}}    
      \label{fig:mean and std of net14}
  \end{subfigure}
  \hfill
  \begin{subfigure}[b]{0.475\linewidth}  
      \centering 
      \includegraphics[width=\textwidth]{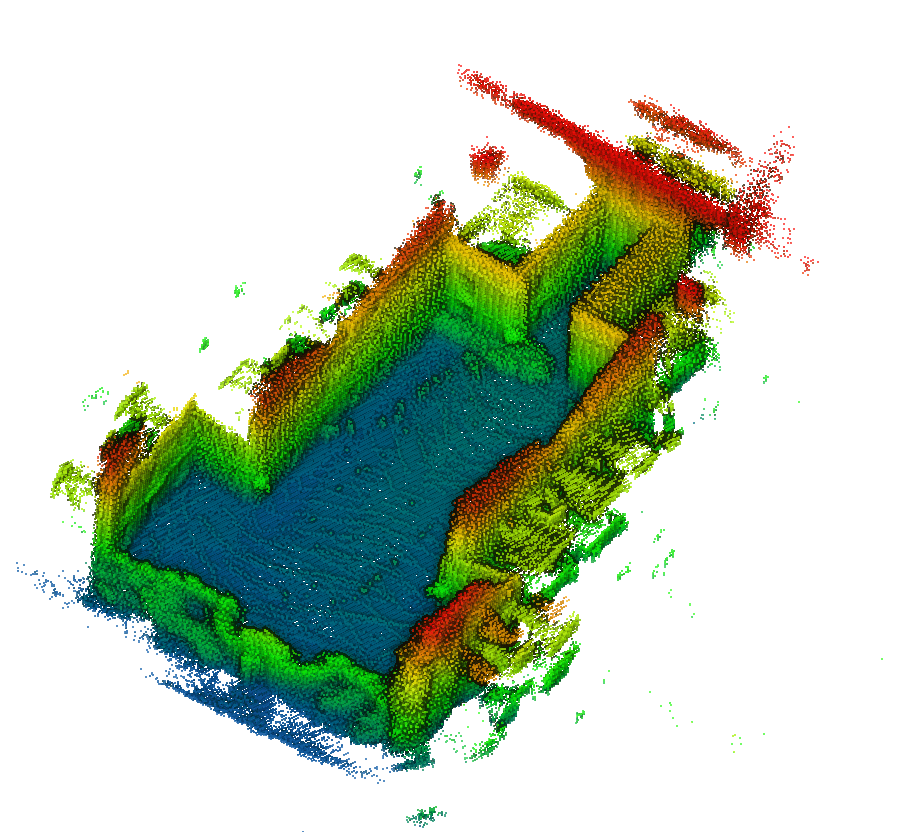}
      \caption[]%
      {{\small Registered LIDAR scans}}    
      \label{fig:mean and std of net24}
  \end{subfigure}
  \vskip\baselineskip
  \begin{subfigure}[b]{0.475\linewidth}   
      \centering 
      \includegraphics[width=\textwidth]{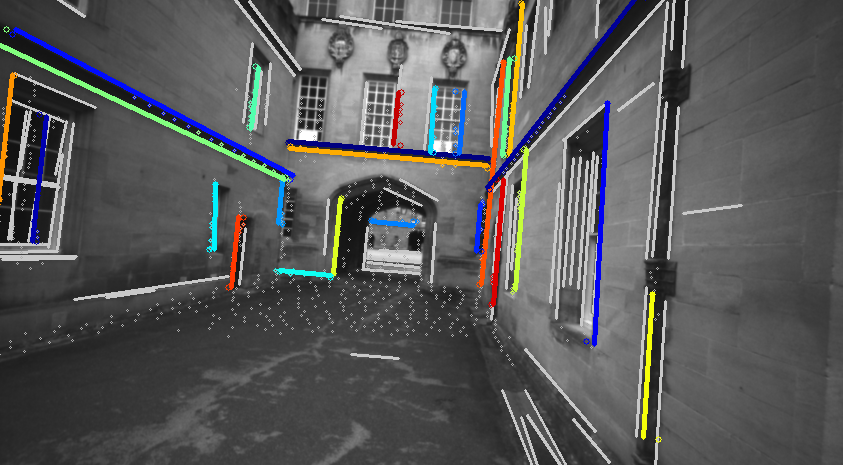}
      \vfill
      \caption[]%
      {{\small Line extraction}}    
      \label{fig:mean and std of net34}
  \end{subfigure}
  \hfill
  \begin{subfigure}[b]{0.475\linewidth}   
      \centering 
      \includegraphics[width=\textwidth]{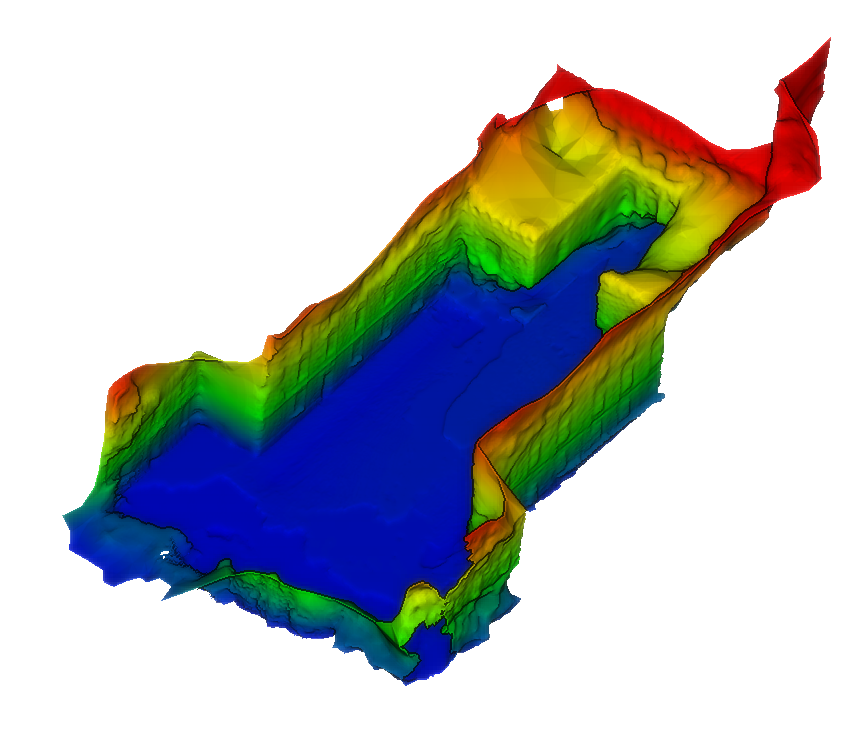}
      \caption[]%
      {{\small Final 3D reconstruction of the scene}}    
      \label{fig:mean and std of net44}
  \end{subfigure}
  \caption[ The average and standard deviation of critical parameters ]
  {\small Our approach, based on lines and points, tightly couples LIDAR scans and monocular images to reconstruct a 3D scene. }
  \label{fig:intro_figure}
\end{figure}
Recently, the fusion of camera imagery and LIDAR scans for 3D reconstruction has been explored as a way to overcome the 
respective challenges of these sensing modalities~\cite{Zhen},~\cite{Zhen2019},~\cite{ZimoLi}. In this approach, camera imagery captures texture, color, and fine details while LIDAR allows accurate range information to be captured that 
is completely independent of visual complexity (although at a lower resolution). This allows it to
fill in gaps where visual feature detection fails.
Indeed, the different natures of these sensors make them complementary, but also makes their joint use complex. \emph{Semantic} or \emph{geometric} approaches aim to enable processing data from these two different sensor modalities in a higher-level space, leveraging the underlying structure of the environment and reducing the complexity that stems from the modality gap.
We focus our work on \emph{lines} as they are very simple geometric primitives, inexpensive to compute and can be easily found in structured environments. 
Finding correspondences between 2D lines across different images, however, is a difficult task owing to the depth ambiguity in monocular images. Indeed, two segments whose endpoints are located on the same epipolar line will be reprojected at the same location in images even if their 3D coordinates
 differ significantly. Therefore, we opt for a 3D line representation that incorporates data from a range sensor, namely LIDAR, along with that from monocular images. 
These correspondences are then exploited to derive a cost function for bundle adjustment which is then optimized to increase the accuracy of the initial pose estimates. 
The optimized lines and the registered LIDAR scans are then incorporated into a LIDAR-Camera Multi-View Stereo (MVS) pipeline. Figure~\ref{fig:pipeline} presents an overview of our workflow.

\begin{figure*}[!h]
  \centering
  \includegraphics[width=\textwidth]{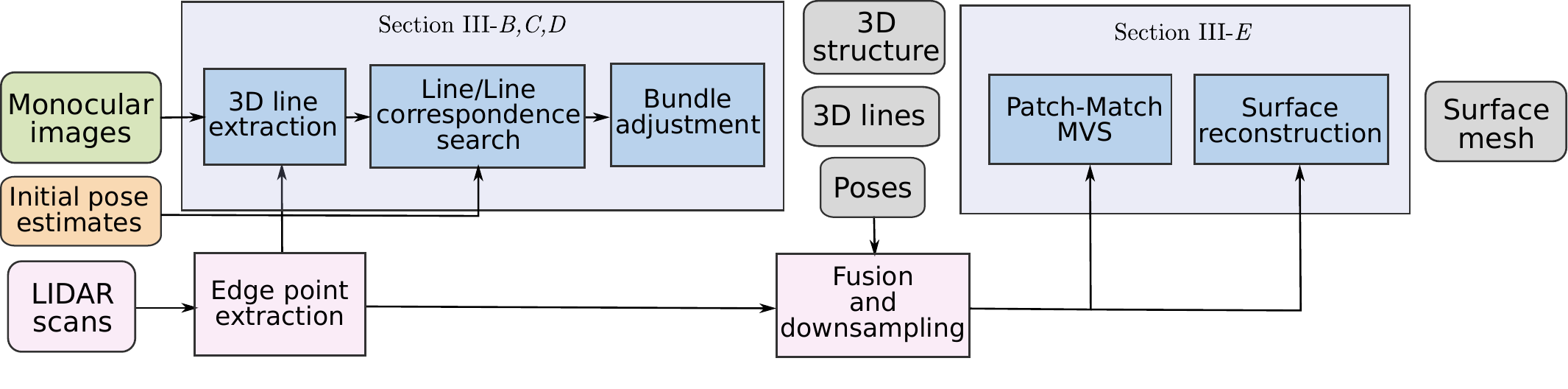}
  \caption{Overview of our LIDAR-camera reconstruction pipeline}
  \label{fig:pipeline}
\end{figure*}

The main contributions of this paper are the following:
\begin{itemize}
  \item We present a novel end-to-end Structure from Motion algorithm that exploits joint LIDAR/camera observations to derive 3D line segments for localization as well as reconstruction.
  \item We experimentally validate our approach by testing it on a public dataset.
\end{itemize}
The remainder of the paper is organized as follows: in section II, we discuss related work in the field of LIDAR-camera fusion for dense reconstruction and SfM using geometric constraints,
 namely line segments. In section III we present our approach to build a surface mesh based on the co-registered line segments, LIDAR scans and camera images. 
 In section IV, we present qualitative and quantitative results of our system applied to public datasets. Finally section V presents concluding remarks and directions for future work.
\section{Related Work}
\subsection{Geometric primitives for bundle adjustment}
State-of-the-art SfM methods such as OpenMVG~\cite{Moulon2017} and Colmap~\cite{schoenberger2016sfm} achieve impressive results with monocular images. 
The use of geometric primitives and more generally, high-level features, has been shown to allow for more efficient computation, although at the cost of more complex algebraic derivation. \cite{LineBundleAdjustmentBartoli} provides a theoretical framework for the use of lines in a bundle adjustment and 
derives an SfM pipeline that is entirely based on lines. Along the same lines, ~\cite{pointlesssSfM} incorporates 2D curves in the bundle adjustment formulation. Moreover, several implementations of 
Simultaneous Localization and Mapping ({SLAM}) approaches have been reported that are based in part or entirely on lines or other geometric primitives (for example {PL-SLAM}~\cite{pumarola2017plslam}). ~\cite{infinitePlanes}  proposes a SLAM algorithm that is based on planes extracted from an RGB-D camera. 
Similarly,~\cite{Rosinol19icra} employs \emph{structural regularities}, such as planes to improve a visual-inertial odometry pipeline.  
~\cite{Yu2020} uses a prior LIDAR map from which 3D lines are extracted and further used to find correspondences with 2D lines extracted from videos. Our approach, while also using line correspondences to enhance a bundle adjustment, does not require the use of a prior LIDAR map and directly creates 3D segments using the LIDAR data along with the camera data. 
\subsection{Lidar-camera fusion}
The robustness of SfM pipelines, however, can be compromised in low-visibility environments (e.g. poor lighting, lack of reliable visual features, etc.) or if the field of view of the camera is narrow.  Numerous recent {SLAM} systems including~\cite{VLOAM},~\cite{LIMO} and ~\cite{SparseVisualLidarSlam} make use of a LIDAR along with a camera to overcome this problem. 
However, the camera data in these implementations is only used to improve localization, or texture a map that is built with LIDAR scans. Conversely, the LIDAR may only be used to add depth information to images. 
\cite{ZimoLi} is, to our knowledge, the first work that leverages camera and LIDAR data for dense reconstruction and extends a Multi-View Stereo (MVS) pipeline to utilize LIDAR points during depth map estimation and surface reconstruction. Although related to our work, it initially assumes that poses are accurately known in order to \emph{consistently} fuse data from those two sensors.  
Other approaches~\cite{Zhen}, \cite{Zhen2019} introduce a probabilistic framework for a joint optimization based on LIDAR patches and points from stereo images.
Our work, albeit similar to the recent work from~\cite{Huang2020LidarMonocularVO}, which also presents a framework dealing with lines and exploiting information from LIDAR scans, is directly based on correspondences between 3D line segments. Moreover, since we focus on a Structure from Motion approach, we do not just optimize on a set of recent images but on all images, making robust data association more challenging. To address this, our approach leverages~\cite{CLEAR}, a multi-view data association framework to check the consistency of line segment data associations.
\section{Methods}
\subsection{Overview}
Before delving into the description of the algorithmic pipeline, we introduce some notation and definitions that will be used throughout the paper.\\
We consider a set of $N$ images $\left\{I_i\right\}_{i\in\mathcal{I}}$ and LIDAR scans $\left\{S_i\right\}_{i\in\mathcal{I}}$. The set of camera poses is denoted by $\mathcal{P}$ and the camera pose corresponding to the $i$-th view is $\mathbf{P_i}$.\\
The projection function $\pi^i$ projects a point from the world frame to the $i$-th image plane such that if $\textbf{X}$ is a 3D point in homogeneous coordinates, and $x$ its 2D projection onto the image $I_i$, 
$$x = \pi^i\left(\mathbf{X}\right) = \mathbf{K}\mathbf{P_i}\mathbf{X}$$ 
where $\mathbf{K}$ represents the intrinsic parameters matrix.\\
A 3D infinite line, can be represented by its Pl\"ucker coordinates $\mathbf{L}\in\mathbb{R}^6$ -- or equivalently its Pl\"ucker matrix --, and a finite line, $\mathbf{l}$, by its two 3D endpoints.
With a slight modification of notation, we define the 3D line segment's projection as being the projection of its two endpoints from the world frame to the image $i$ frame such that $\pi^i\left(\mathbf{l}\right) = (\pi^i\left(^s\mathbf{l}\right), \pi^i\left(^e\mathbf{l}\right))$ where $^s\mathbf{l}$ represents the starting point of the 3D line and $^e\mathbf{l}$ its end point. In the case of an infinite line $\pi^i(\mathbf{L})$ represents the three coefficients of the 2D line's equation corresponding to the projection of the 3D infinite line on the $i$-th image.
$\mathcal{X}$ represents the set of 3D landmarks used for SfM and the set of observations associated to a 3D landmark $\mathbf{X}\in\mathcal{X}$ is referred to as $\mathcal{O}_\mathbf{X}$.\\
We assume that we have initial estimates of the camera poses that can be obtained for example with a monocular visual-inertial {SLAM} system like VINS~\cite{vins}. 
Those priors will help fuse the LIDAR data by (1) providing relative rotations and translations between the camera viewpoints and (2) providing the scale of the scene.\\
As illustrated in Fig.2, our pipeline includes two components. The first is an enhanced Structure from Motion component which processes on 3D lines, previously detected by a line detection
module. The line detection module processes data from the LIDAR and the camera to extract and merge 3D lines in the different views. Using correspondence search we are then able to
create clusters of lines to yield an additional error term in the bundle adjustment's cost function. Optimization over the new cost function yields new pose estimates. 
The second component leverages the new pose estimates and the 3D segments to reconstruct a surface mesh that is based on combined information from images and LIDAR scans.   
\subsection{Line detection}
The line detection module aims to extract 3D lines that are observed by the camera as well as by the LIDAR. 

\begin{figure}
  \centering
  \includegraphics[width=\linewidth]{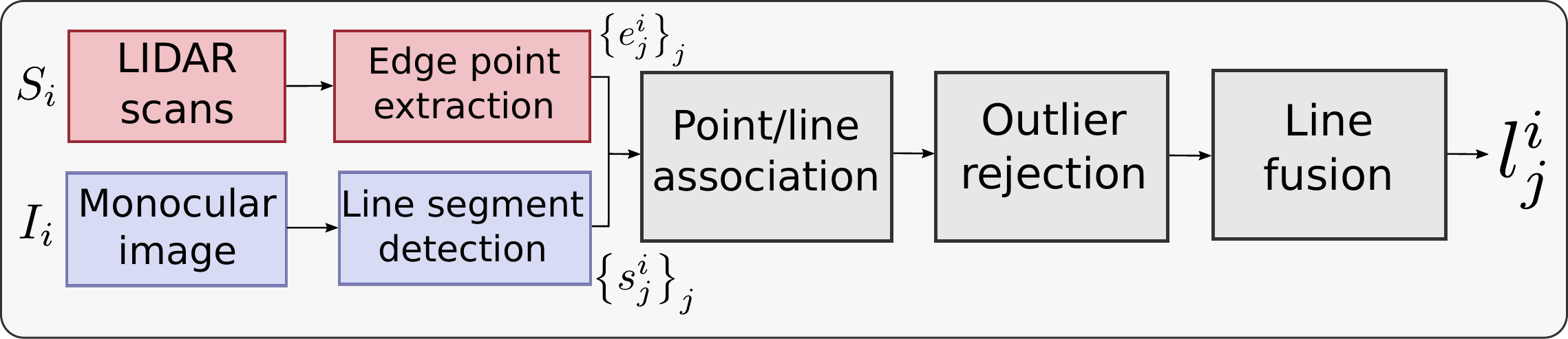}
  \caption{Overview of the camera/LIDAR 3D segment detection pipeline}
  \label{fig:step1}
\end{figure}

We begin by identifying potential 2D line segments in images which will constitute the basis to derive the subsequent 3D lines. Line Segment Detector (LSD)~\cite{GromponeVonGioi2012} 
is a widely used method that accurately detects \emph{line segments} based on \emph{line support regions}. We use this detector to output a set of line segments $\left\{s_j^i\right\}_j$
visible in image $I_i$. Then, we augment the 2D information with range information from the LIDAR (\textit{i.e.} incorporating a third dimension) and thus to be able to precisely locate the line segment in 3D.
Associating the correct LIDAR points to the detected 2D line segments is crucial to the performance of our algorithm. Unlike the system described in \cite{Huang2020LidarMonocularVO}, which directly associates the closest LIDAR point to the detected 2D line segments, we aim to find LIDAR points that \textit{effectively} correspond to the observation of the same line. Since LIDAR scans do not provide any texture information, we discard lines corresponding to flat surfaces (\emph{i.e.} only corresponding to changes in texture), and instead focus the LIDAR/camera association on \emph{edges} (\emph{i.e.} lines that correspond to changes in the local surface normal).
Hence, in addition to the 2D line segments, we retrieve edge points from LIDAR scans based on a smoothness score as described in~\cite{EnglotLegoLOAM}.
The resulting edge points are then matched to the 2D segments based on their pixel distance. For each 2D segment $s_i^j$ we have now associated a set of LIDAR edge points
$\left\{e^i_j\right\}_j$. By construction, we know that the selected LIDAR points correspond to a 3D line. However, an ambiguity may remain as to the specific line to which those
points correspond. To resolve any ambiguity and ensure that those edge points correspond to the same line that matches the detected 2D segment, we run RANSAC on the set of edge points. Only inliers are kept and we compare the reprojected 3D line to the 2D segment outputted by LSD. Finally, to avoid multiple instances of the same \textit{actual} line segment in an image, we also fuse 3D segments having similar directions and which are located on the same infinite line.
This procedure is illustrated Fig.\ref{fig:step1}.
\subsection{3D/3D line correspondences}
Once 3D segments have been detected across the set of images, we look for segments that represent different views of the same actual segment \emph{via} 3D/3D correspondences.
Let us consider a 3D segment $\mathbf{l}$ detected in the $i$-th image. We first reproject the current line segment on all the images using its 3D information and only keep the views in which the segment is visible. We then iterate over the set of segments detected in those views and compute a similarity score, as described below. 
Algorithm 1 provides a detailed view of the matching code: 

\begin{algorithm}
  \caption{Find all matches for 3D line segment $l$}
  \begin{algorithmic}
  \REQUIRE $t_{\theta}, t_{dP}, t_{dO}, t_{LBD}$ //matching thresholds 
  \STATE $\text{matches}\leftarrow \left\{\right\}$
  \FOR{each image $I_j$}
  \IF{$l$ is not in $I_j$'s field of view} 
  \CONTINUE
  \ENDIF
  \STATE $s_{mini} \leftarrow \infty$, $l_{mini} \leftarrow \oslash$
  \FOR{each segment $l{'}$ detected in image $I_j$}
  \STATE $a\leftarrow angle(l,l',j)$
  \STATE $b\leftarrow distPixel(l,l',j)$
  \STATE $c\leftarrow distOrtho(l,l',j)$
  \STATE $d\leftarrow distFeat(l,l',j)$

  \IF{$a < t_{\theta}$ and $b<t_{dP}$ and $c<t_{dO}$ and $d<t_{LBD}$}
  \STATE $s \leftarrow a/t_{\theta}+b/t_{dP}+c/t_{dO}+d/t_{LBD}$
  \IF{$s < s_{mini}$}
  \STATE $s_{mini} \leftarrow s$, $l_{mini}\leftarrow l{'}$
  \ENDIF
  \ENDIF
  \ENDFOR
  \STATE matches $\leftarrow\text{matches}\cup \left\{l_{mini}\right\}$
  \ENDFOR
  \end{algorithmic}
  \end{algorithm}
The similarity score $s$ takes into account:
\begin{itemize}
  \item The angle between the reprojected 3D segments;
  \item The pixel distance between the projected endpoints;
  \item The "orthogonal distance" defined as the norm of the vector $\frac{1}{2}(\mathbf{^sl}+\mathbf{^el}) - \frac{1}{2}(\mathbf{^sl}'+\mathbf{^el'})$ projected perpendicularly to the 3D segment's direction;
  \item In addition to the three criteria described above, we also use line band descriptors (LBD)~\cite{LBD} which provide additional information about the local appearance of the line.
\end{itemize}
Each of these four components is weighted by four hyperparameters $\alpha_{\theta}, \alpha_{\text{dist}}, \alpha_{overlap}, \alpha_{LBD}$ that control the relative importance of the different distances. 

  \begin{figure}
  \centering
  \includegraphics[width=\linewidth]{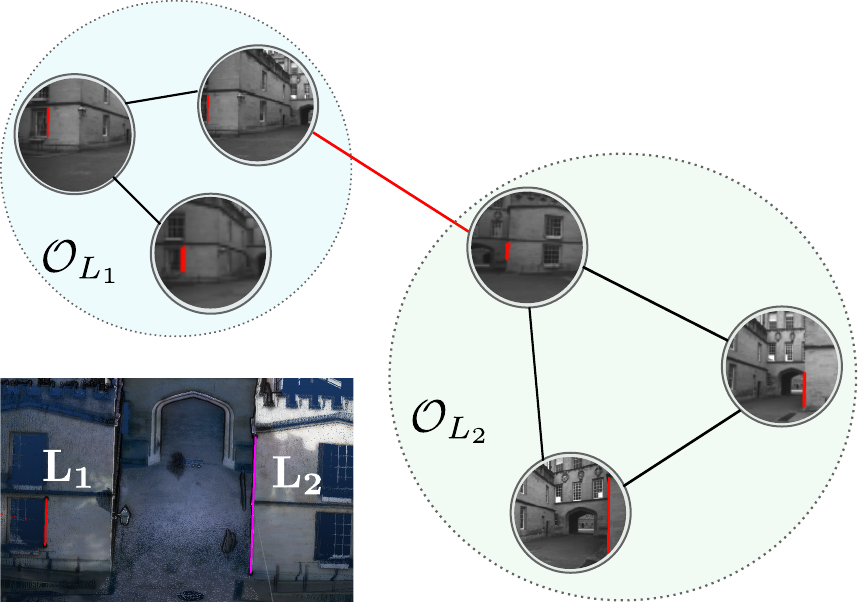}
  \caption{Pairwise associations (edges) are made following the matching procedure previously described.
  In red an inconsistent data association filtered out by CLEAR~\cite{CLEAR}  . 
  $\mathcal{O}_{\mathbf{L_1}}$ and $\mathcal{O}_{\mathbf{L_2}}$ are sets of 3D line segments $\mathbf{l^i_j}$ from different views but representing the same segment
   (here resp. $\mathbf{L_1}$ and $\mathbf{L_2}$) whose accurate location is as yet unknown.}
  \label{fig:da}
\end{figure}

Algorithm 1 describes the procedure to compute the set of matches for a single line segment. We repeat the procedure for all line segments which eventually yields a set of pairwise associations between line segments. However, some of these associations may not be consistent accross different views, where two line segments from the same view visually representing different actual segments can be connected in the data association graph. To ensure the consistency of the pairwise associations computed during the previous step and to mitigate potentially false data association, we use {CLEAR}~\cite{CLEAR}.\\
Based on the filtered pairwise data associations, we infer which line observations $\mathbf{l^i_j}$ correspond to the same \emph{actual} line and build up a set of lines $\mathcal{L} = \left\{\mathbf{L_1}, ..., \mathbf{L_m} |  m\in\mathbb{N}\right\}$ to which we attach a set $\left\{\mathcal{O}_{\mathbf{L_j}}\right\}$ of line observations where we denote by $\mathcal{O}_{L_j}$ the set of observations of line $\mathbf{L_j}$. Because we do not know the exact equation of these lines, we initialize their Pl\"ucker vector with an initial rough estimate based on the 3D information provided by their associated observations $\mathcal{O}_{\mathbf{L_j}}$. Lines are further optimized as described in the following paragraph.
This process is summarized Fig.\ref{fig:da}.
\subsection{Line bundle adjustment}
The $M$ 3D lines instances $\mathbf{L_1}, ..., \mathbf{L_M}$ inferred at the previous step, and their observations allow us to extend the typical  bundle adjustment formulation to take into account lines' reprojection errors.
Optimizing over finite line segments is often non-trivial because the optimized endpoints may end up in occluded areas. Instead, we represent lines in $\mathcal{L}$ as infinite lines during the optimization step.
\\
The following paragraphs detail the different error terms involved in the cost function that we seek to minimize.\\
1) Camera observations\\
We define the \emph{point reprojection error} of a 3D feature $\mathbf{X_j}\in\mathcal{X}$ as the sum over its observations ($x^i_k\in\mathcal{O}_{\mathbf{X_j}}$) of the pixel distance between its reprojection onto the image plane and its associated observation for that image:
\begin{equation}
  \label{eqn:errorPoint}
  e_R\left(\mathbf{X_j}\right) = \sum\limits_{i\in\mathcal{I}}\sum\limits_{x^i_k\in\mathcal{O}_{\mathbf{X_j}}}{||\pi^i\left(\mathbf{X_j}\right) - x^i_k||^2}
\end{equation}

2) Line reprojection:\\
The \emph{line reprojection error} of a line $\mathbf{L_j}\in\mathcal{L}$ is defined as being the sum over all its observations of the distances between the observation and the reprojection of line $\mathbf{L_j}$ on the $i$-th image \emph{i.e.} $\pi^i\left(\mathbf{L_j}\right)$. 
Analogously to \cite{pumarola2017plslam}, if we consider an observation $\mathbf{l^i_k}$ detected in the $i$-th image of the line $\mathbf{L_j}$, the error will be defined as the sum of the 2D distances between the projected starting point  $\pi^i(\mathbf{^sl^i_k})$ and the reprojected infinite line $\pi^i(\mathbf{L_k})$, and between the projected end point of $\mathbf{^el^i_j}$ and $\pi^i(\mathbf{L_j})$.
\begin{equation}
  \label{eqn:errorLine}
  e_{L}\left(\mathbf{L_j}\right) = \sum\limits_{i\in\mathcal{I}}\sum\limits_{\mathbf{l^i_k}\in\mathcal{O}_{\mathbf{L_j}}} ||\mathbf{^sl^i_k}, \pi^i(\mathbf{L_j})||_{2D} + ||\mathbf{^el^i_k}, \pi^i(\mathbf{L_j})||_{2D} 
\end{equation}
Based on the error terms~\eqref{eqn:errorPoint} and~\eqref{eqn:errorLine} introduced above, the line bundle adjustment cost function can be written as follows:
\begin{equation}
  \label{eqn:optimization}
  \mathcal{P}^*, \mathcal{X}^*, \mathcal{L}^* = \argmin_{\mathcal{P},\mathcal{X},\mathcal{L}} \lambda_R\sum_{\mathbf{X_j}\in\mathcal{X}}e_R\left(\mathbf{X_j}\right) + \lambda_{L}\sum_{\mathbf{L_k}\in\mathcal{L}}{e_L\left(\mathbf{L_k}\right)}
\end{equation}
where $\lambda_L$ and $\lambda_R$ are coefficients chosen to change the relative importance of line factors in the optimization and $\mathcal{P}^*, \mathcal{X}^*, \mathcal{L}^*$ represent respectively the optimized 3D poses, the optimized 3D landmark locations 
and the optimized infinite line equations. The optimization of~\eqref{eqn:optimization}, implemented with the library Ceres~\cite{ceres}, is performed using the Levenberg-Marquart algorithm. We resort to the orthonormal representation~\cite{orthoLines} for lines during the optimization process.
This is a minimal representation, using only four parameters, as opposed to Pl\"ucker coordinates which require six parameters. As a last step for this module, we transform the infinite lines back 
to line segments.

\subsection{Lidar-enhanced multi-view stereo}
We now present the last step of our pipeline, which exploits further the detected line segments along with LIDAR scans and camera images. Typically an MVS pipeline includes three steps: depth map computation, depth map fusion and mesh reconstruction, as described \cref{fig:pipeline}. Our contribution strictly focuses on the first and last steps.

Provided with the poses computed in the previous step, we register all of the LIDAR point clouds onto a single point cloud $S_{fused}$ and downsample it using a voxel grid. 
As in~\cite{ZimoLi}, we keep track of the number of LIDAR points that have been clustered in the same voxel.  This metric represents a measure of the visibility and relative
importance of each voxel for subsequent steps. Depth map computation follows the approach described in~\cite{ZimoLi}, in that we rely on 
LIDAR scans to initialize the depth maps. We slightly modify their approach, however, to reduce potential occlusions:
Given the registered LIDAR scan $S_{fused}$ we initialize each depth map's pixel using the depth of a visual feature, if available.  Otherwise, we initialize with the range provided by the LIDAR.
Instead of handling occlusions at a pixel's scale (by taking the closest to the camera LIDAR measurement in case of multiple LIDAR measurements available for one pixel) as in~\cite{ZimoLi},
we extend the verification to a neighborhood. We check the consistency of the pixel's depth with respect to the average depth in the neighborhood. We keep the LIDAR measurement only
if the depth is consistent with the local average.
After computing depth maps, we fuse them as in~\cite{depthFusion}. The last step consists in producing a surface mesh based on the fused depth maps and the registered LIDAR scans.
We extend the approach described in~\cite{ZimoLi} which ingeniously combines points from the registered LIDAR scans to points from the dense visual feature cloud. In addition to using this combination we make use of the 
previously detected line segments. We sample points on each line segment and add them as vertices in the Delaunay triangulation.
\section{Results}
We test our approach on a state-of-the-art public dataset, \emph{The Newer College Dataset}~\cite{ramezani2020newer}, which provides sequences of images and scans of structured (buildings) environments.
\subsection{Newer college dataset}
\begin{figure}
  \centering
    \includegraphics[width=\linewidth]{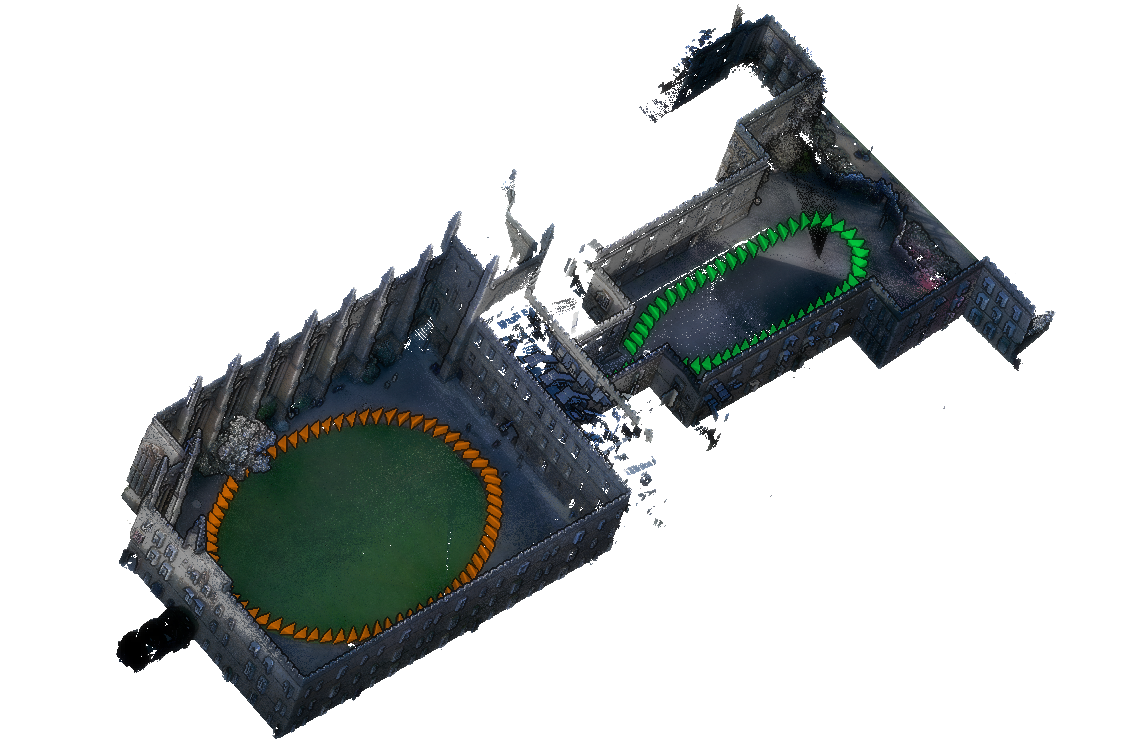}
\caption{Ground truth trajectory. The first sequence is displayed in \emph{green}, the second in \emph{orange}.}
\label{fig:figure_traj}
\end{figure}

The Newer College Dataset consists of different sequences recorded using a handheld rig carried through the New College in Oxford. An Ouster OS-1 64 LIDAR and an Intel Real-Sense D435i are mounted on the rig. In addition, ground truth point cloud data is provided for reconstruction using a Leika BLK360 survey scanner, as well as localization data for the Ouster LIDAR that was derived using the Iterative Closest Point algorithm.
We test our approach on two sequences extracted from the dataset with each containing approximately sixty images. Intrinsic calibration of the camera as well as extrinsic calibration between the camera and the LIDAR are provided. Figure~\ref{fig:figure_traj} shows the camera's positions in our sequence and a 3D view of the dataset.
We generate noisy initial pose estimates by adding random translational ($\Sigma_T=0.5m$) and rotational noise ($\Sigma_R=5^{\circ}$) to the ground truth trajectory, but this could also be data provided by an IMU for example. 
\begin{figure}
  \centering
  \includegraphics[width=\linewidth]{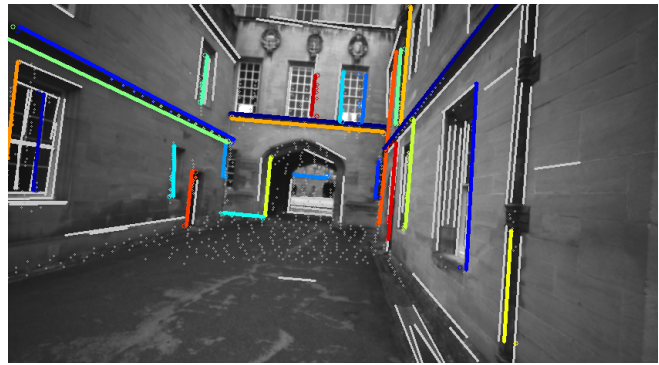}
  \caption{Each circle represents a detected edge point from the LIDAR scan, lines are detected using a Line Segment detector. The color represents an association between a 2D line segment and edge points. Gray color represents edges and lines that have not been associated}
  \label{cref:lines_newer_college}
\end{figure}
\subsection{Evaluation}
\begin{figure}
  \centering
  \begin{subfigure}[b]{0.475\linewidth}
    \centering
    \includegraphics[width=\textwidth]{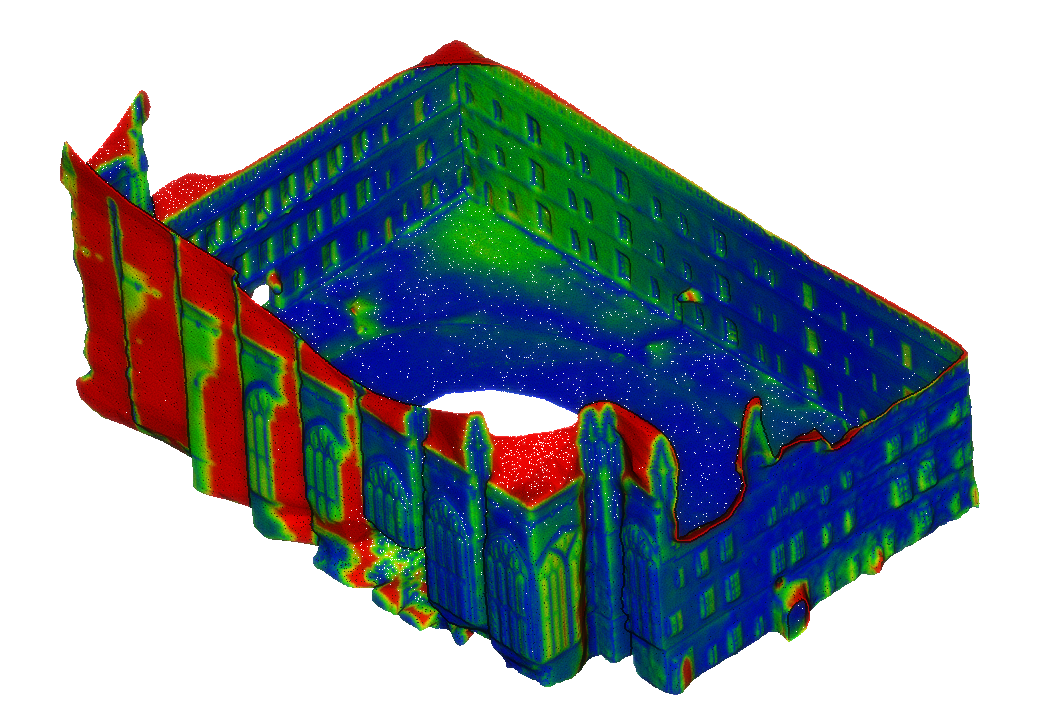}
    \caption[Network2]%
    {{\small Photogrammetry~\cite{Moulon2017},\cite{openmvs2020}\label{mvg_mvs}}}    
  \end{subfigure}
  \hfill
  \begin{subfigure}[b]{0.475\linewidth}  
    \centering 
    \includegraphics[width=\textwidth]{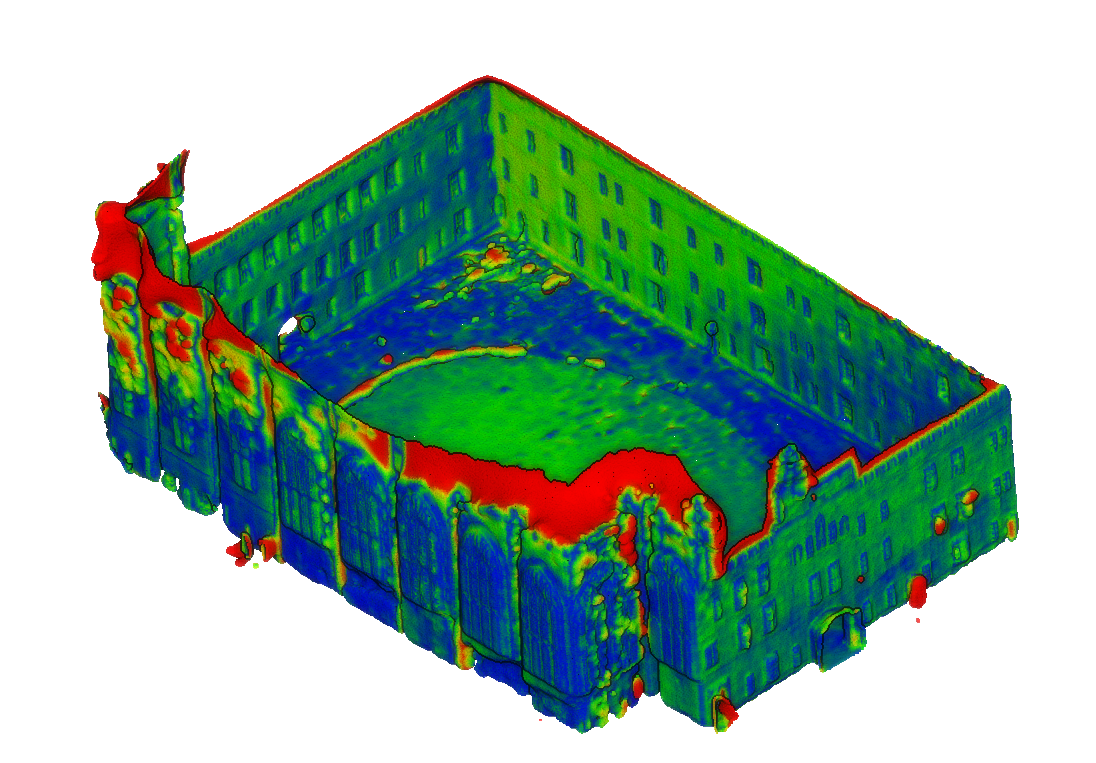}
    \caption[]%
    {{\small LIDAR only~\cite{poissonRecon}\label{lidar}}}    
  \end{subfigure}
  \vskip\baselineskip
  \begin{subfigure}[b]{0.475\linewidth}   
    \centering 
    \includegraphics[width=\textwidth]{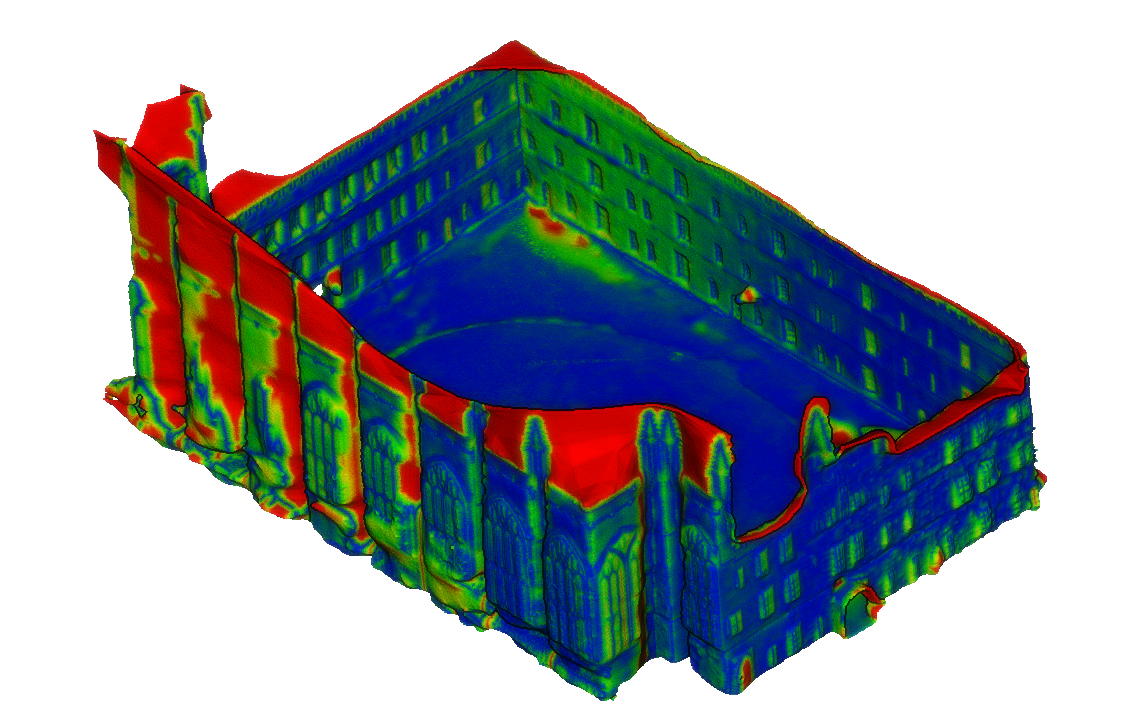}
    \vfill
    \caption[]%
    {{\small Z.Li et al.~\cite{ZimoLi} \label{zli}}}    
  \end{subfigure}
  \hfill
  \begin{subfigure}[b]{0.475\linewidth}   
    \centering 
    \includegraphics[width=\textwidth]{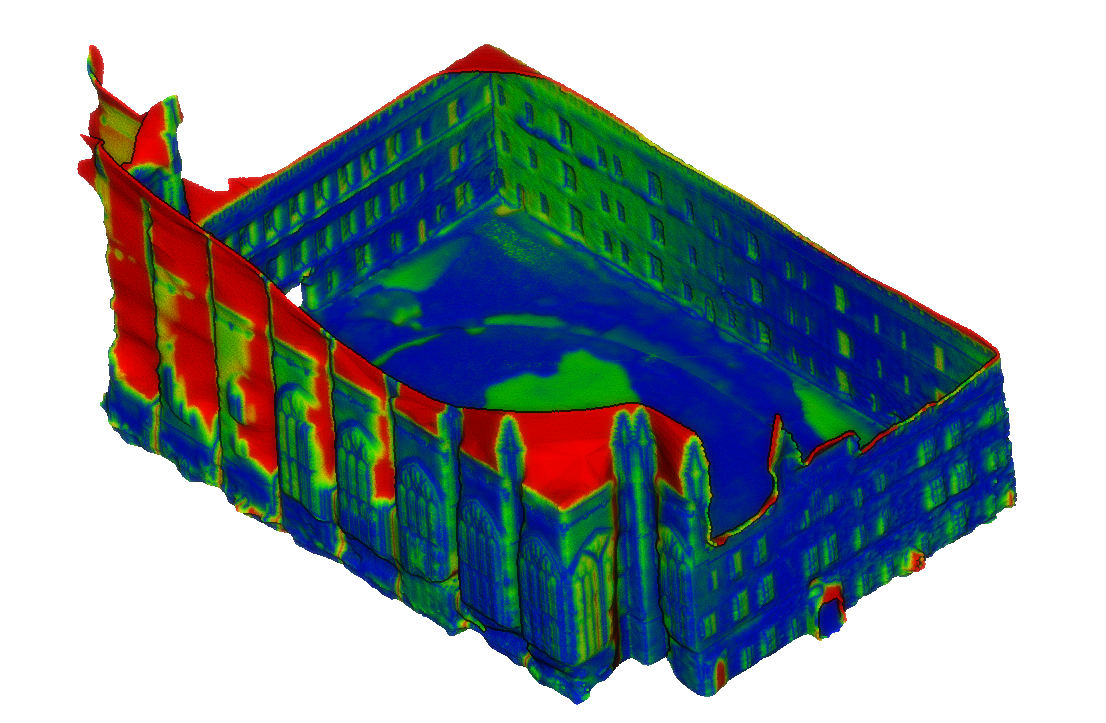}
    \caption[]%
    {{\small Ours \label{ours}}}    
  \end{subfigure}
  \begin{subfigure}[b]{\linewidth}
    \centering
    \includegraphics[width=\textwidth]{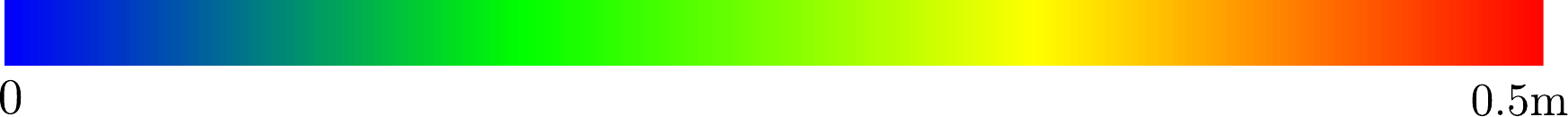}
  \end{subfigure}
  \caption{Surface meshes obtained with different pipelines on sequence \#2. False colors represent the distance of the mesh to ground truth point cloud from blue (close to 0) to red (more than $0.5$m)}
  \label{fig:comparison_meshes}
\end{figure}

All experiments were run on an Intel Core i7-8850 CPU with 16GB of RAM.
We first evaluate our line-bundle adjustment against a state-of-the art SfM approach~\cite{Moulon2017} using \emph{average localization error}, a metric which characterizes the accuracy of the estimated camera positions.
\begin{table}
\begin{center}
\begin{tabular}{ccccc}\toprule
  \multirow{2}{*}{Metric} & \multicolumn{2}{c}{Newer College \#1} & \multicolumn{2}{c}{Newer College \#2}\\
  & SfM~\cite{Moulon2017} & Ours &  SfM~\cite{Moulon2017} & Ours\\
  \midrule
  Avg. localization error (m.)& 0.142 & \textbf{0.140} & 0.130 &\textbf{0.129}\\
\end{tabular}
\caption{Numerical evaluation of our approach on \emph{The Newer College Dataset}. Best result shown in \textbf{bold}.}
\label{sfm_results}
\end{center}
\end{table}
Since SfM does not compute poses at the metric scale we estimate the unknown scale factor along with the $SE(3)$ transformation matrix to the world frame using Umeyama alignment with
the initial noisy poses as an input. Our approach, in constrast, directly utilizes data at metric scale provided by the LIDAR and therefore directly reconstructs the scene at the exact scale. The results in ~\cref{sfm_results} show that using lines jointly with visual points yields slightly better results than the SfM pipeline for both sequences that were tested.
We attribute this improvement to the greater stability that line segments provide relative to point features.  This stability derives from the fact that they may remain visible in a more diverse set of views than point features, which may be matched in only a handful of images.
\\
We then evaluate our mesh reconstruction using \emph{precision} and \emph{recall} as metrics characterizing both the accuracy of the
resulting mesh and its completeness, along with the \emph{F-score} that aggregates both results. As in~\cite{ZimoLi}, because our ground truth 
reconstruction is provided as a point cloud, we first need to transform our mesh to a point cloud and remove occluded areas before comparing it to ground truth.
We set the distance threshold to $0.5m$ to compute precision and recall. The main evaluation results, obtained and visualized using the software \emph{CloudCompare}, are summarized in \cref{table:results}.
Our algorithm is compared to three other pipelines: a photogrammetry pipeline based on the open source libraries OpenMVG~\cite{Moulon2017} and OpenMVS~\cite{openmvs2020}, a LIDAR-only surface reconstruction algorithm~\cite{poissonRecon} based on the ground truth poses, and the approach described in~\cite{ZimoLi}.
Illustrations of the final meshes obtained by the different approaches are shown~\cref{fig:comparison_meshes}.

\begin{table}
  \begin{center}
  \begin{tabular}{cccccc}\toprule
    Metric & Photogrammetry  & LIDAR only & Z. Li et al.  & Ours\\ 
     & \cite{Moulon2017},~\cite{openmvs2020}& \cite{poissonRecon}& \cite{ZimoLi} & \\
  \midrule
    Precision  &\textbf{0.85}/\textbf{0.97} &0.77/0.90 & 0.82/0.92& 0.81/0.93\\
    Recall  &0.45/0.84 &\textbf{0.75}/\textbf{0.97} & 0.69/0.93 & 0.69/0.93\\
    F-Score &0.59/0.90 &\textbf{0.76}/\textbf{0.93} & 0.75/\textbf{0.93} & 0.74/\textbf{0.93}\\
  \end{tabular}
\end{center}
\caption{Reconstruction results for the \emph{Newer College Dataset sequence \#1/sequence \#2}. Best result shown in \textbf{bold}}
\label{table:results}
\end{table}
It is evident that photogrammetry results suffer from a lack of camera coverage in some areas and sometimes does not properly render featureless parts of the environment, including some walls and parts of the ground. In contrast, LIDAR-based approaches cover a significantly larger area, but at the cost of precision. As expected, the best overall performance is realized with the LIDAR-only approach, which relies on the highly accurate ground truth poses from the Newer College Dataset. 
Notably, our approach, which does not require accurate initial poses unlike the one in~\cite{ZimoLi} (as previously discussed), exhibits an F-score that is better than that of photogrammetry and comparable to the LIDAR-only case for both test sequences. Our approach is also noticeably more precise in areas where 3D line segments have indeed been detected. Finally, we note that our approach is more applicable to larger scenes since it does not rely on having access to a dense LIDAR point cloud of the environment for depth map initialization.  
\section{Conclusion and future work}
We have presented a novel pipeline for LIDAR and camera-based surface reconstruction that is based on 3D line segments: geometric primitives 
that are computationally inexpensive to obtain, lightweight, and which embed information from both sensors. Starting from noisy pose and line estimates and leveraging state-of-the-art data association algorithms,
we are able to exploit optimized line segments and poses to realize an improved a multi-view stereo pipeline.
The synthesis of accurate visual features from the camera, the dense -- but less accurate -- LIDAR point clouds, 
and the rigid structure information provided by line segments enables a higher quality surface reconstruction than a camera-only approach and comparable performance to a gold standard LIDAR-only 3D reconstruction that relies on highly accurate ground truth pose estimates.

One of the critical assumptions of this work is that the environment be of a structured nature. This requirement makes it appropriate in urban settings. Future work may explore other primitives that could be used jointly with lines to improve performance and relax the assumption of a highly structured environment. \cite{manifoldThinStructMVS}, for example, presents an interesting approach to leverage 2D curves during reconstruction to allow recovery of finer details.
\addtolength{\textheight}{-4cm}   

\section{Acknowledgements}
The authors wish to acknowledge Stéphane Vannuffelen, Sepand Ossia, and Kevin Doherty for insightful discussions and guidance.
\printbibliography

\end{document}